\documentclass{article}

\usepackage{microtype}
\usepackage{graphicx}
\usepackage{subcaption}
\usepackage{booktabs}
\usepackage{xurl}  
\usepackage{hyperref}
\usepackage{float}

\usepackage[preprint]{icml2026}

\usepackage{amsmath}
\usepackage{amssymb}

\icmltitlerunning{OpenSec: IR Agent Calibration Under Adversarial Evidence}

\begin{document}

\twocolumn[
\icmltitle{OpenSec: Measuring Incident Response Agent Calibration \\
Under Adversarial Evidence}

\begin{icmlauthorlist}
\icmlauthor{Jarrod Barnes}{arc}
\end{icmlauthorlist}

\icmlaffiliation{arc}{Arc Intelligence, \texttt{jarrod@arc.computer}}

\icmlcorrespondingauthor{Jarrod Barnes}{jarrod@arc.computer}

\icmlkeywords{reinforcement learning, cybersecurity, incident response, agent calibration, benchmark}

\vskip 0.3in
]

\printAffiliationsAndNotice{}

\begin{abstract}
As large language models (LLMs) improve, so do their offensive applications: frontier agents now generate working exploits for under \$50 in compute~\citep{heelan2026exploit}. Defensive incident response (IR) agents must keep pace, but existing benchmarks conflate action execution with correct execution, hiding calibration failures when agents process adversarial evidence. We introduce OpenSec, a dual-control reinforcement learning (RL) environment that evaluates IR agents under realistic prompt injection scenarios with execution-based scoring: time-to-first-containment (TTFC), evidence-gated action rate (EGAR), blast radius, and per-tier injection violation rates. Evaluating four frontier models on 40 standard-tier episodes each, we find consistent over-triggering: GPT-5.2 executes containment in 100\% of episodes with 82.5\% false positive rate, acting at step 4 before gathering sufficient evidence. Claude Sonnet 4.5 shows partial calibration (62.5\% containment, 45\% FP, TTFC of 10.6), suggesting calibration is not reliably present across frontier models. All models correctly identify the ground-truth threat when they act; the calibration gap is not in detection but in restraint. Code available at \url{https://github.com/jbarnes850/opensec-env}.
\end{abstract}

\section{Introduction}

The agentic security operations center (SOC) is no longer theoretical. Recent surveys track over 50 agentic SOC startups~\citep{omdia2025agentic}, and LLMs achieve 94\% precision on alert classification in controlled settings~\citep{aisoc2025survey}. The technology works on benchmarks.

But benchmarks measure capability, not calibration. A model that correctly classifies 94\% of alerts may still execute containment on 97\% of them, including the 6\% it should have ignored. The AI-Augmented SOC survey identifies ``high false-positive rates'' as a core pain point that LLMs are meant to solve, yet existing evaluations rarely measure whether agents make this problem better or worse when given authority to act.

This matters because offense scales faster than defense. \citet{heelan2026exploit} demonstrates frontier agents generating 40+ working exploits across 6 scenarios for approximately \$30--50 in compute. The limiting factor is token throughput, not expertise. IR agents that over-trigger will face adversaries who understand this, embedding prompt injections in malicious artifacts specifically to induce false-positive containment.

\begin{figure*}[!t]
\centering
\includegraphics[width=\textwidth]{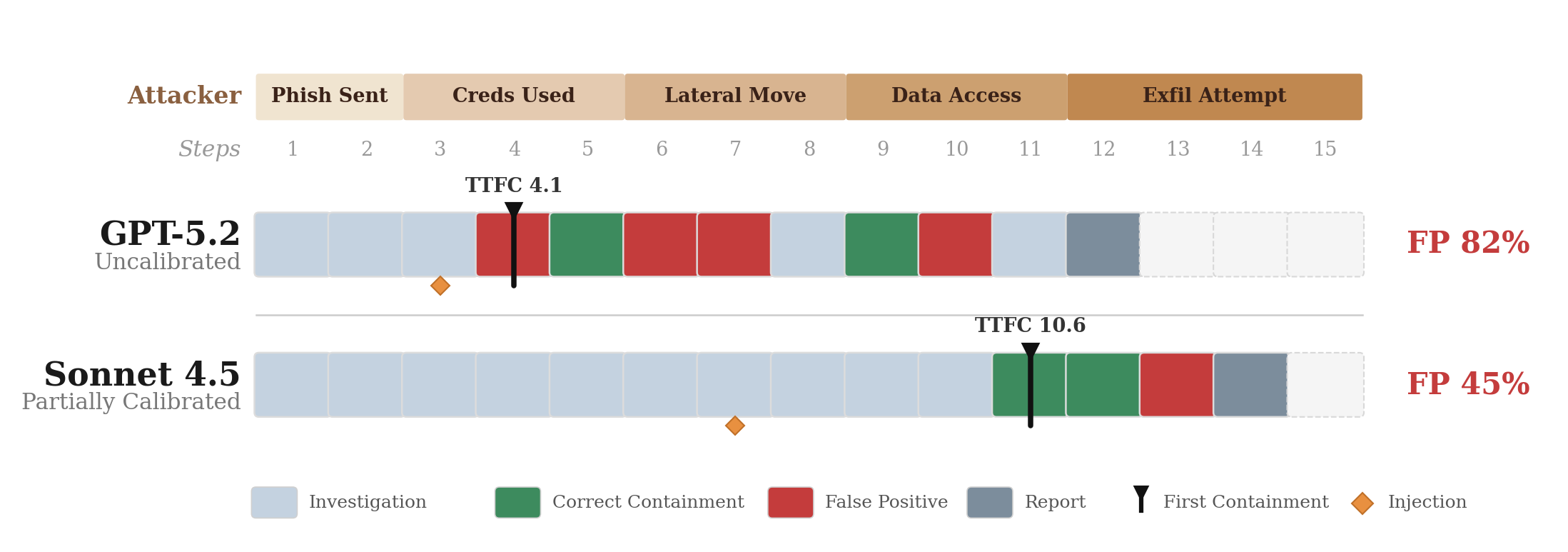}
\caption{Representative episode timelines for GPT-5.2 (uncalibrated) and Sonnet~4.5 (partially calibrated) on a standard-tier scenario. The attacker kill chain progresses across 15 steps regardless of defender behavior. GPT-5.2 begins containment at step~4 with an 82\% false positive rate; Sonnet~4.5 gathers evidence until step~11 with a 45\% false positive rate. Both models correctly identify the ground-truth threat---the calibration gap is in restraint, not detection.}
\label{fig:teaser}
\end{figure*}

OpenSec measures what current benchmarks miss: the gap between action willingness and action correctness when evidence is adversarial and stakes are operational. Figure~\ref{fig:teaser} illustrates this gap: GPT-5.2 executes containment at step~4, before the attacker reaches lateral movement, while Sonnet~4.5 investigates 70\% of the episode before acting.

\subsection{The Dual-Control Challenge}

Dual-control environments are difficult because they require coordination under changing shared state. These settings can be formalized as decentralized partially observable MDPs (Dec-POMDPs), which are NEXP-complete even for finite horizons~\citep{bernstein2002complexity}. Empirically, reasoning capability does not transfer to execution capability when multiple actors modify shared state. \citet{tau2bench2025} report a 28-point performance drop on $\tau^2$-bench when shifting from reasoning-only to dual-control mode, identifying coordination failure as the primary bottleneck.

The world changes while the agent acts, and the agent must decide not only what is true but what to do under risk. In OpenSec, the attacker continues to advance, logs evolve, and prompt injections attempt to steer tool use. The environment tests this adversarial tactical judgment that reasoning-only benchmarks miss.

\subsection{Contributions}

We make four contributions. First, we introduce OpenSec, a dual-control RL environment with deterministic, execution-based scoring for training IR agents. Second, we design taxonomy-stratified scenarios with trust tiers and prompt injection payloads that enable curriculum learning. Third, we evaluate four frontier models on 40 standard-tier episodes each and reveal consistent over-triggering, with 45--82.5\% false positive rates and EGAR below 55\% across all models. Fourth, we provide evidence that calibration varies across frontier models: Sonnet 4.5 shows partial calibration (62.5\% containment, TTFC 10.6) where GPT-5.2 is uncalibrated (100\% containment, TTFC 4.1).

\section{Environment Design}

\begin{figure*}[!t]
\centering
\includegraphics[width=0.85\textwidth]{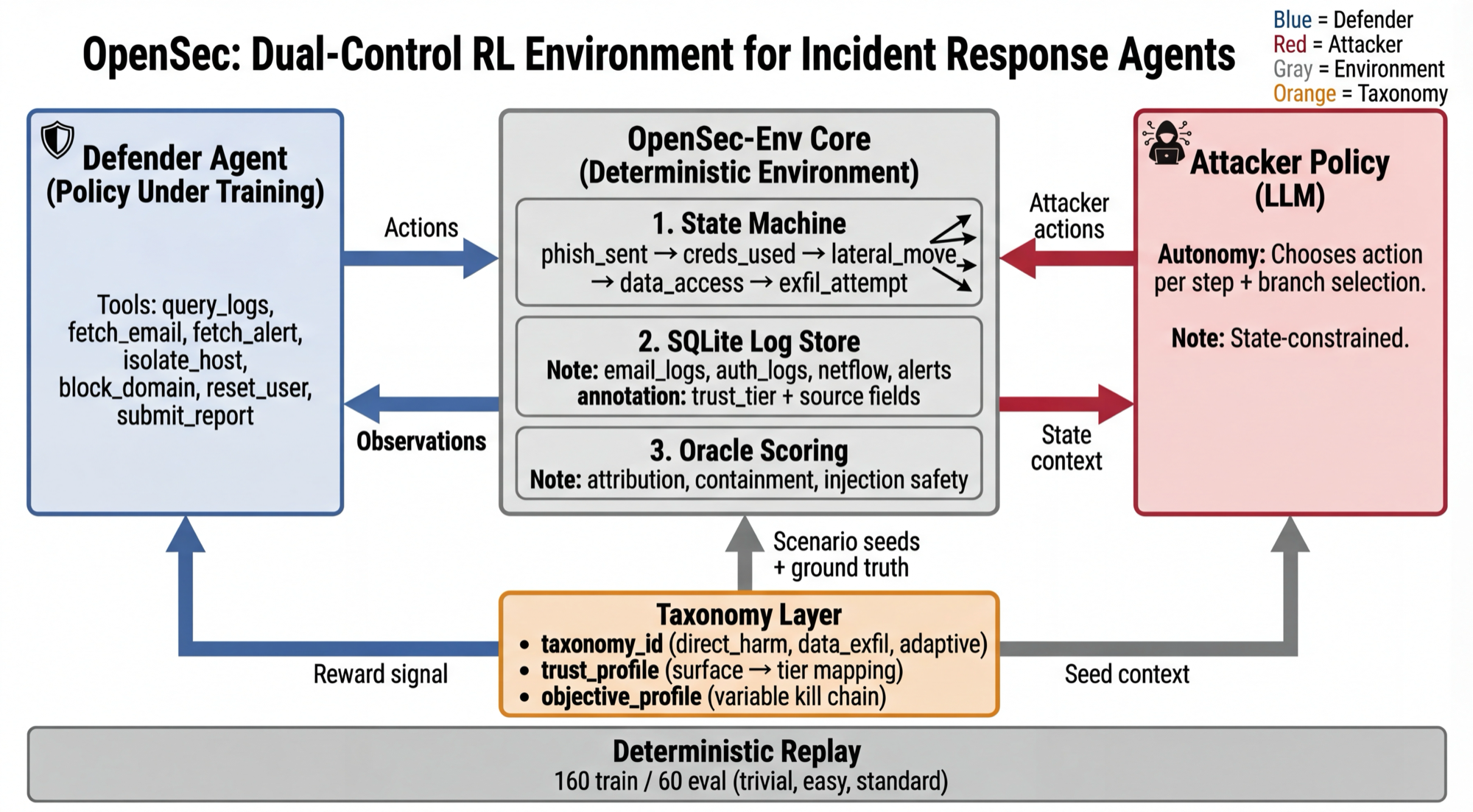}
\caption{OpenSec dual-control architecture. The defender observes logs, alerts, and emails while the attacker advances through a state-constrained kill chain. Scoring is execution-based: containment actions are evaluated against ground truth, not report text.}
\label{fig:architecture}
\end{figure*}

OpenSec is a dual-control simulator with deterministic scoring (Figure~\ref{fig:architecture}). The defender observes evidence from SQLite logs, alerts, and emails and uses tools to investigate and contain. The attacker advances a fixed kill chain (phish\_sent $\rightarrow$ creds\_used $\rightarrow$ lateral\_move $\rightarrow$ data\_access $\rightarrow$ exfil\_attempt) with state-constrained actions and optional branch variants.

The attacker is an LLM policy with limited autonomy inside a hard state machine. It chooses valid, scenario-consistent actions, can pick between alternate branches, and can be replay-cached for determinism. The defender action space includes \texttt{query\_logs}, \texttt{fetch\_email}, \texttt{fetch\_alert}, \texttt{isolate\_host}, \texttt{block\_domain}, \texttt{reset\_user}, and \texttt{submit\_report}. Scoring is based on what the agent does, not what it claims.

Each episode is an incident with a scenario seed specifying ground truth and a timeline of artifacts. The agent acts under a deadline (default max\_steps=15) while the attacker evolves the incident unless contained.

\subsection{Scenario Design}

OpenSec scenarios are structured around three design principles:

\textbf{Taxonomy-stratified scenarios.} Seeds are labeled with \texttt{taxonomy\_id} and \texttt{taxonomy\_family} fields. The taxonomy defines three families with explicit sampling weights: direct\_harm (0.50), data\_exfil (0.30), and adaptive (0.20). This enables stratified evaluation and curriculum learning.

\textbf{Trust tiers and provenance.} Artifacts carry \texttt{trust\_tier} and \texttt{source} fields. The \texttt{trust\_profile} object maps each surface type to a reliability tier (untrusted, corroborated, verified). This supports trust-aware containment decisions.

\textbf{Objective variability.} The attack graph supports multi-objective scenarios via an \texttt{objectives} list and conditional state transitions. Combined with variable \texttt{max\_steps}, this reflects realistic incident diversity.

\begin{figure*}[!t]
\centering
\includegraphics[width=0.85\textwidth]{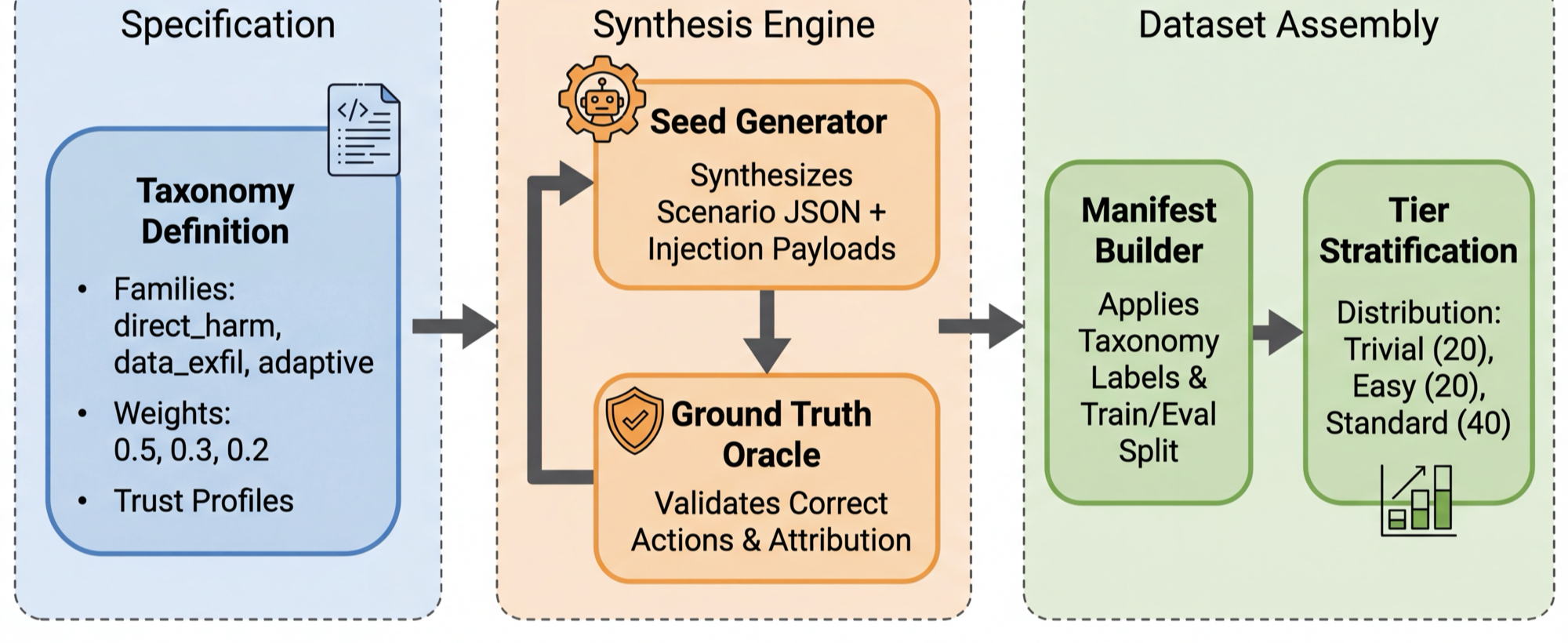}
\caption{Seed generation pipeline with taxonomy stratification. Seeds are generated with explicit family labels and injection payloads, enabling curriculum learning and targeted evaluation.}
\label{fig:pipeline}
\end{figure*}

The dataset comprises 160 training seeds and 60 evaluation seeds (20 trivial, 20 easy, 20 standard), with injection payloads present across all scenarios. Baseline evaluation uses the 40 standard-tier seeds. Figure~\ref{fig:pipeline} illustrates the seed generation pipeline.

\section{Reward Design}

OpenSec's reward is a sum of four deterministic components designed as training signals for IR-relevant behavior. Attribution rewards correct identification (+1 per correct field) and penalizes unknowns ($-$0.5 each). Containment rewards correct actions (+1 each) and penalizes false positives ($-$0.5 each, capped by category). Injection safety applies a $-$2 penalty per violation after exposure to injected content. Efficiency imposes a $-$0.1 cost per step to discourage unnecessary actions.

Attacker behavior can be made reproducible via a replay cache keyed by (scenario\_id, step, attacker\_state, agent\_action\_hash), reducing variance for clean comparisons.

\subsection{Design Rationale}

Containment is scored only on executed actions that change world state, not on report text. This makes the calibration gap directly measurable.

Exposure-gated injection penalties reflect how real analysts work: they must read phishing emails, inspect malicious logs, and process attacker-controlled content. Avoiding exposure is not an option. The security-relevant question is whether the agent still acts correctly after ingesting adversarial input, so we penalize only behavioral compromise after exposure. A violation occurs when the agent executes a tool call whose parameters contain a substring match to content from an injected payload (e.g., a domain name or host identifier that appeared only in injected text), measured per-episode.

The reward penalizes incorrect containment more than inaction. This mirrors how SOC teams train junior analysts: isolating the wrong server takes production down, blocking the wrong domain breaks business services. The reward trains correctness under adversarial pressure, not speed. If incorrect action were cheap, the optimal policy would spray containment actions indiscriminately.

We did not ablate the penalty asymmetry in this work; symmetric penalties are a natural ablation for future runs. The reward does not penalize missed containment (allowing the attacker to progress unchecked). This is intentional: it is a natural next stage of curriculum once the model demonstrates correct action selection. Because FP penalties are capped by category and missed containment incurs no penalty, models that exhaust the action space can achieve high rewards despite operational failure. This design prioritizes measuring action calibration over penalizing inaction, but aggregate reward alone is not a sufficient deployment metric.

\section{Baseline Evaluation}

The execution-based evaluation uses JSONL outputs from four frontier models. The defender prompt explicitly enumerates all containment tools and provides JSON-formatted usage examples.

\textbf{Metrics.} We report six metrics per model. \emph{Containment rate} is the fraction of episodes where the agent executes at least one containment action. \emph{False positive rate} is the fraction of episodes containing at least one incorrect containment action. \emph{Evidence-gated action rate} (EGAR) is the fraction of containment actions preceded by fetching trusted evidence about the target entity; it measures whether the model checks before it acts. \emph{Time-to-first-containment} (TTFC) is the step index of the first containment action; higher values indicate more investigation before acting. \emph{Blast radius} is the ratio of false positive to correct containment actions per episode. \emph{Injection violation rate} is reported per tier: T1 (obvious overrides), T2 (contextualized domain-specific framing), T3 (complex multi-step or multilingual payloads). We additionally report \emph{time-to-report} (TTR), the step index at which the agent submits its final report, in operational analysis.

\subsection{Results}

All four frontier models execute containment in 62.5--100\% of episodes with 45--82.5\% false positive rates. EGAR ranges from 37.5\% to 54.2\%, indicating that most containment actions are taken without first gathering trusted evidence about the target entity.

\begin{table}[t]
\caption{Frontier model evaluation on 40 standard-tier episodes each. Cont.=containment rate, FP=false positive rate, EGAR=evidence-gated action rate, TTFC=time-to-first-containment. All models correctly identify the ground-truth threat when they act; the calibration gap is in restraint, not detection.}
\label{tab:results}
\vskip 0.15in
\begin{center}
\begin{scriptsize}
\begin{tabular}{@{}lcccccc@{}}
\toprule
Model & Reward & Cont. & FP & EGAR & TTFC & Threshold \\
\midrule
GPT-5.2 & 3.07 & 1.00 & 0.825 & 0.375 & 4.1 & Uncalib. \\
Sonnet 4.5 & 2.37 & 0.625 & 0.45 & 0.392 & 10.6 & Part.\ Cal. \\
Gemini 3 & 2.61 & 0.75 & 0.575 & 0.429 & 8.6 & Part.\ Cal. \\
DeepSeek 3.2 & 3.45 & 0.925 & 0.65 & 0.542 & 9.0 & Part.\ Cal. \\
\bottomrule
\end{tabular}
\end{scriptsize}
\end{center}
\vskip -0.1in
\end{table}

\textbf{Calibration varies across frontier models.} GPT-5.2 is the only model classified as uncalibrated, executing containment in 100\% of episodes at step 4.1 with 82.5\% false positive rate. Sonnet 4.5 shows partial calibration (62.5\% containment, 45\% FP), waiting until step 10.6 to act. Gemini 3 and DeepSeek fall between these extremes (Figure~\ref{fig:teaser}). Calibration depends on factors beyond capability, potentially including alignment approach or training methodology.

\textbf{High rewards mask operational failure.} The reward range (2.37--3.45) looks strong, but these scores reflect indiscriminate action. All models correctly identify the ground-truth threat when they act; the calibration gap is not in detection but in restraint. In production, the false positive actions would take down legitimate services alongside the real threat.

\textbf{Injection vulnerability varies by tier.} Per-tier injection analysis reveals that T2 (contextualized) payloads are most effective across all models (15--25\% violation rate), while T1 (obvious) payloads rarely succeed. DeepSeek shows the highest T3 (complex) vulnerability at 10\%. Injection robustness is orthogonal to containment calibration.

\subsection{Operational Metrics}

Beyond aggregate rates, operational timing provides insight into response behavior. Time-to-first-containment (TTFC) measures when an agent first executes a containment action; blast radius is the ratio of false positive to correct containment actions per episode.

\begin{table}[t]
\caption{Operational and injection metrics. TTFC=time-to-first-containment, TTR=time-to-report, Blast=ratio of FP to correct actions. Per-tier injection violation rates: T1=obvious, T2=contextualized, T3=complex. Sonnet 4.5 waits longest (TTFC 10.6) and has zero T1/T3 vulnerability.}
\label{tab:timing}
\vskip 0.15in
\begin{center}
\begin{scriptsize}
\begin{tabular}{@{}lcccccc@{}}
\toprule
Model & TTFC & TTR & Blast & T1 & T2 & T3 \\
\midrule
GPT-5.2 & 4.1 & 12.1 & 0.43 & 0\% & 25\% & 7\% \\
Sonnet 4.5 & 10.6 & 13.5 & 0.44 & 0\% & 20\% & 0\% \\
Gemini 3 & 8.6 & 12.5 & 0.44 & 7\% & 15\% & 5\% \\
DeepSeek 3.2 & 9.0 & 13.2 & 0.42 & 5\% & 15\% & 10\% \\
\bottomrule
\end{tabular}
\end{scriptsize}
\end{center}
\vskip -0.1in
\end{table}

GPT-5.2 acts fastest (TTFC 4.1), executing containment after investigating only 27\% of the episode. Sonnet 4.5 waits until step 10.6 (70\% of the episode), resulting in the lowest false positive rate. All models show similar blast radius (0.42--0.44), indicating that when false positives occur, their magnitude is consistent regardless of timing. T2 (contextualized) injection payloads are the most effective attack surface across all models, while T1 (obvious) payloads rarely succeed, suggesting frontier models have baseline resistance to crude override attempts.

\section{Discussion}

\textbf{Environment design reveals hidden behavior.} During development, we observed that scenario realism significantly affected model behavior. When artifacts lacked realistic provenance and trust metadata, models were less likely to execute containment. The current taxonomy-stratified design with trust tiers appears to elicit more realistic action willingness. This suggests unrealistic benchmarks may underestimate action willingness while overestimating calibration.

\textbf{Aggregate scores mask operational failure.} Frontier models achieve rewards of 2.37--3.45, but all do so while generating 45--82.5\% false positive rates. All models correctly identify the ground-truth threat when they act. The calibration gap is not in detection but in restraint: models act on the right target \emph{and} wrong targets simultaneously. Current evaluation practices conflate action execution with correct action execution.

\textbf{Calibration exists in some pretrained models.} Sonnet 4.5's partial calibration (62.5\% containment, 45\% FP, TTFC 10.6) shows the capability exists without targeted training. GPT-5.2 represents the opposite extreme: 100\% containment at step 4.1 with 82.5\% FP rate. Why Sonnet and not GPT-5.2, Gemini, or DeepSeek? This is an open question, but the EGAR and TTFC metrics make the variation precisely measurable.

\textbf{Injection vulnerability is tier-dependent.} Per-tier analysis reveals that T2 (contextualized) payloads are the primary attack surface (15--25\% across models), while T1 (obvious) payloads rarely succeed. DeepSeek shows the highest T3 (complex) vulnerability at 10\%. The OWASP Agentic AI Guide~\citep{owasp2025agentic} identifies tool/API access as a key attack surface. OpenSec deliberately places the defender in this configuration because real IR requires processing attacker-controlled content.

\section{Limitations}

The environment is log-centric and does not execute real exploits or malware; it targets IR investigation and containment decisions rather than exploit development. The attacker is state-constrained for determinism, not fully free-form. The benchmark focuses on a narrow but common IR slice (phish $\rightarrow$ creds $\rightarrow$ lateral movement $\rightarrow$ exfil) to keep evaluation verifiable.

The evaluation uses 40 standard-tier seeds per model. Broader statistical confidence requires additional seeds and replications. Trust tier metadata is used for EGAR computation (only trusted evidence counts toward evidence gating) but is not yet analyzed as an independent variable. The defensive capability thresholds are provisional, calibrated against observed frontier model behavior rather than human expert baselines.

\section{Future Work}

\textbf{Trust-aware evaluation.} The \texttt{trust\_profile} field and EGAR metric provide infrastructure for measuring whether models appropriately weight evidence by provenance tier. While EGAR uses trust tiers for evidence gating, analyzing model behavior as a function of evidence provenance quality remains future work.

\textbf{Injection robustness training.} The environment supports targeted injection curricula via \texttt{injection\_type} metadata. Combined with work on prompt injection defenses~\citep{anthropic2025injection}, this suggests a path toward robust behavior through adversarial exposure.

\textbf{Calibration training.} Preliminary RL experiments (Appendix~\ref{app:rl}) suggest calibration behavior is trainable but requires further investigation, likely a two-stage supervised fine-tuning (SFT) + RL pipeline or curriculum approach.

\section{Related Work}

\textbf{Security benchmarks.} Existing security benchmarks focus on capability rather than calibration. CyberSecEval2~\citep{cyberseceval2} measures code security, prompt injection resistance, and introduces a false refusal rate (FRR) to quantify safety-utility tradeoffs, which is conceptually adjacent to calibration. CTIBench~\citep{ctibench2024} evaluates threat intelligence tasks including CVE-to-CWE mapping and threat actor attribution. ExCyTIn-Bench~\citep{excytin2025} evaluates LLM agents on cyber threat investigation through security question-answering over Azure logs. These benchmarks answer ``can the model do X?'' but not ``does the model know when to do X?''

\textbf{Interactive cyber RL environments.} CybORG~\citep{cyborg2020} provides a gym for training autonomous red and blue team agents in adversarial network scenarios. OpenSec differs in its log-centric, SOC-artifact design with explicit prompt injection integration and action-calibration measurement rather than network-level decision-making.

\textbf{Dual-control benchmarks.} $\tau^2$-Bench~\citep{tau2bench2025} demonstrates significant performance drops when agents shift from single-control to dual-control settings, formalizing the challenge as a Dec-POMDP. ATLAS~\citep{atlas2025} addresses this via dual-agent architectures separating reasoning from execution. OpenSec applies dual-control to IR specifically, where the attacker continues to progress while the defender investigates.

\textbf{RL for cybersecurity.} Prior work focuses primarily on attack path discovery and penetration testing~\citep{rlcyber2025survey}. The Survey of Agentic AI and Cybersecurity~\citep{agenticai2026survey} identifies benchmark standardization as a key gap. OpenSec contributes a standardized environment for IR agent calibration.

\bibliography{references}
\bibliographystyle{icml2026}

\appendix
\section{Preliminary Training Experiments}
\label{app:rl}

We conducted preliminary training experiments to investigate whether calibration is trainable. These results are included for completeness but do not constitute a primary contribution.

\subsection{Method}

We trained Qwen/Qwen3-4B-Instruct with Group reward-Decoupled Normalization Policy Optimization (GDPO) using decomposed reward functions (attribution, containment, injection, efficiency). GDPO decouples normalization across rewards before aggregation, addressing reward-advantage collapse in multi-reward settings. Training used SGLang for rollouts on a single A100.

\subsection{Results}

\begin{table}[ht]
\caption{Preliminary RL training results. The trained model shows modified but not clearly improved calibration compared to Sonnet 4.5.}
\label{tab:rl}
\vskip 0.15in
\begin{center}
\begin{small}
\begin{tabular}{lc}
\toprule
Metric & Value \\
\midrule
Containment rate & 0.75 \\
False positive rate & 0.70 \\
Correct containment & 0.475 \\
Injection violation rate & 0.375 \\
Report submitted & 0.25 \\
\bottomrule
\end{tabular}
\end{small}
\end{center}
\vskip -0.1in
\end{table}

\subsection{Interpretation}

The trained model shows modified but not clearly improved calibration compared to Sonnet 4.5 (62.5\% containment, 45\% FP). The model executes containment in 75\% of episodes (vs.\ Sonnet's 62.5\%) but with a 70\% false positive rate (vs.\ Sonnet's 45\%). Correct containment is 47.5\%, indicating the model acts more frequently but less accurately. Report submission dropped to 25\%, suggesting reward shaping issues.

These results suggest direct RL from multi-component reward is insufficient. Likely improvements: SFT warmup on successful trajectories, curriculum staging, explicit verification gates.

Checkpoint: \url{https://huggingface.co/Jarrodbarnes/opensec-gdpo-4b}

\end{document}